\documentclass[conference]{IEEEtran}
\usepackage{hyperref}
\hypersetup{
    colorlinks=true,
    linkcolor=black, 
    citecolor=black,    
    urlcolor=blue    
}
\usepackage{amsmath,amssymb,amsfonts,cleveref}
\usepackage{algorithmic}
\usepackage{graphicx}
\usepackage{float}
\usepackage{textcomp}
\usepackage{xcolor}
\def\BibTeX{{\rm B\kern-.05em{\sc i\kern-.025em b}\kern-.08em
    T\kern-.1667em\lower.7ex\hbox{E}\kern-.125emX}}

\begin{document}

\title{GRASP: Municipal Budget AI Chatbots for Enhancing Civic Engagement}

\author{\IEEEauthorblockN{Jerry Xu\textsuperscript{1}}
\IEEEauthorblockA{\textit{Lexington High School} \\
Lexington, MA \\
jerry.max.xu@gmail.com}
\and
\IEEEauthorblockN{Justin Wang\textsuperscript{1}}
\IEEEauthorblockA{\textit{Lexington High School} \\
Lexington, MA \\
justinwangma@gmail.com}
\and
\IEEEauthorblockN{Joley Leung\textsuperscript{2}}
\IEEEauthorblockA{\textit{Lexington High School} \\
Lexington, MA \\
Joley.Leung@gmail.com}
\and
\IEEEauthorblockN{Jasmine Gu\textsuperscript{2}}
\IEEEauthorblockA{\textit{Lexington High School} \\
Lexington, MA \\
jasminegu101@gmail.com}
}

\maketitle

\begin{abstract}
There are a growing number of AI applications, but none tailored specifically to help residents answer their questions about municipal budget, a topic most are interested in but few have a solid comprehension of. In this research paper, we propose \textbf{GRASP}, a custom AI chatbot framework which stands for \textbf{G}eneration with \textbf{R}etrieval and \textbf{A}ction \textbf{S}ystem for \textbf{P}rompts. GRASP provides more truthful and grounded responses to user budget queries than traditional information retrieval systems like general Large Language Models (LLMs) or web searches.  These improvements come from the novel combination of a Retrieval-Augmented Generation (RAG) framework (``Generation with Retrieval") and an agentic workflow (``Action System"), as well as prompt engineering techniques, the incorporation of municipal budget domain knowledge, and collaboration with local town officials to ensure response truthfulness. During testing, we found that our GRASP chatbot provided precise and accurate responses for local municipal budget queries 78\% of the time, while GPT-4o and Gemini were only accurate 60\% and 35\% of the time, respectively. GRASP chatbots greatly reduce the time and effort needed for the general public to get an intuitive and correct understanding of their town’s budget, thus fostering greater communal discourse, improving government transparency, and allowing citizens to make more informed decisions.
\end{abstract}


\section{Introduction}

In modern democracies, accessing municipal budget information has become much easier. The main challenge with promoting public awareness of this information now lies in the efficient browsing of the vast amount of information online, and the derivation of comprehensible and accurate summaries of this information.  

Many local governments release annual budget documents that detail the financial plans for the upcoming fiscal year, as well as provide budget information from previous years. However, despite these documents being available to the public, challenges still remain for users wishing to find accurate budget information. For instance, web searches can direct users to unofficial or outdated documents, and in most general large-language models (LLMs) like ChatGPT, their knowledge base is not updated frequently enough to account for document revisions or newly-released documents. It is crucial that citizens are informed based solely on official and up-to-date documents to ensure that the information they are receiving is as accurate as possible.

Even users looking at the latest documents may experience difficulties. For example, lookup in documents (e.g. by ``CTRL+F searching”) uses plaintext search rather than semantic search. Therefore, a user investigating the total budget allocated to their municipal public schools may search for ``total school budget” rather than ``total education budget”, and hence be unable to find the statistic even though the two queries effectively mean the same thing. 

Another complication in accuracy arises when considering the distinction between ``actual” and ``projected” values. For example, if someone wishes to find information on the municipal school budget for 2024, they would likely go to the FY2024 budget document. However, since budget documents are made before the fiscal year in question, the value listed in the FY2024 document is merely a projected value. The true value could be found in the FY2025 document after the 2024 fiscal year had ended and the actual budget total was tallied. This is another common pitfall of both web searches and general LLMs.

These factors contribute to the overall decrease in public understanding of how their local government allocates its budget, highlighting the need for more intuitive and context-aware information retrieval systems.

A chatbot has the potential to resolve those challenges. It is clear that an LLM can process and summarize large amounts of data much more quickly than a human ever can, so the key to creating an ideal informational chatbot is to ensure that its responses are both truthful and grounded (i.e. they are not hallucinations). 

Our research paper makes two key contributions. First, we propose the GRASP framework to improve the accuracy, completeness, and groundedness of information retrieval systems focusing on the domain of municipal budget, thus increasing resident trust. Secondly, we ensure GRASP is highly generalizable for all towns and cities. 

\section{Related Works}

Our GRASP framework encompasses all of the following topics:

\subsection{Overview of Chatbots Powered by Large Language Models (LLMs)}

LLMs, such as GPTs (Generative Pre-trained Transformers), represent the forefront of advancements in AI’s ability to process and understand natural language. These models have provided chatbots with the ability to understand and generate human-like responses \cite{patil2024transformative}. LLMs utilize multiple layers of neural networks, including embedding layers, attention layers, and feed-forward networks, which work together to process input text and generate output content \cite{NIPS2017_3f5ee243}. Additionally, LLMs are pre-trained on vast textual datasets sourced from the internet, encompassing trillions of words. Studies have evaluated how these models can enhance the accuracy and efficiency of information retrieval \cite{10.1145/3605801.3605806}. However, LLMs suffer from hallucinations, moments when the model generates an output that is factually incorrect or fabricated due to the biases or incomplete information present in their training datasets \cite{kirchenbauer2024hallucination}, thus presenting the need for Retrieval Augmented Generation (RAG).

\subsection{Retrieval-Augmented Generation (RAG) – Optimized LLMs}

The concept of retrieval-augmented generation was first introduced by Lewis et al. (2021) \cite{NEURIPS2020_6b493230}. RAG systems introduce a mechanism to retrieve from authoritative external knowledge bases, reducing hallucinations that traditional LLMs suffer from. In addition to hallucinations, traditional LLMs may also contain different versions of a dataset from what is specified as their ``cutoff" date, providing users with misleading and incorrect data \cite{cheng2024dateddatatracingknowledge}. While Fine-Tuning (Dodge et al.,2020) \cite{DBLP:journals/corr/abs-2002-06305} can be used to inject new knowledge into pre-trained LLMs through a series of specialized training \cite{wang2024evaluatingqualityanswersretrievalaugmented}, RAG systems exhibited higher levels of efficiency and accuracy in responses, consistently outperforming fine-tuned models \cite{lakatos2024investigatingperformanceretrievalaugmentedgeneration} \cite{wang2024evaluatingqualityanswersretrievalaugmented}. Because of the various advantages that RAG systems have in retrieving the most up-to-date information, we chose RAG to be the basis of our town budget chatbot.

\subsection{ReAct Agents}

The ReAct (or ReAct agent) paradigm combines an LLM’s abilities to reason and create action plans, proposing that an agent performs tasks via repeated thoughts-action-observation steps. Specifically, the thoughts step allows for the creation and adaptation of action plans, the action step performs tasks such as gathering information from external sources like knowledge bases, and the observation step analyzes the action's output and determines if more thoughts-action-observation steps are necessary to respond to the user (iterative refinement) \cite{yao2022react}. ReAct proved to be a capable tool for question-answering and fact verification, for which it utilized HotpotQA \cite{yang2018hotpotqa} and Fever \cite{thorne2018fever} respectively. Additionally, when tested against imitation and reinforcement learning models, ReAct outperformed both by an absolute success rate of 34 percent and 10 percent respectively. 

\subsection{State of the Art of RAG systems in Government}

In Brazil, 20 percent of the Public Administration's estimated 13 trillion USD annual spending is allocated to public works. However, unfinished work is commonplace for legal reasons, including corruption, negatively impacting the entire region. In 2024, Honório et al. created a chat assistant trained on data from Brazilian public works, aiming to solve this issue. Public works databases provided to the assistant spanned from 2010 to 2019, and this data included databases on Companies' Registers, Status of Public Works Contracts, Contract attributes, and Public Works Attributes. When evaluated, the model proved to be accurate, and when tested by a group of 35 individuals of varying ages and education, the vast majority deemed the assistant's answers clear, helpful, and accurate. By providing residents with a resource to track public works spending,  this chat assistant effectively demonstrated  the role of RAG and AI in civic engagement and civilian understanding of the complexities associated with public works. \cite{honorio2024large}.

\section{GRASP}

\emph{Problem Definition: design an AI tool that is capable of searching municipal budget documents and providing truthful and grounded responses to user queries.}

GRASP uniquely integrates all of the following methods to ensure response accuracy:

\subsection{RAG (Retrieval Augmented Generation) Framework}

The RAG framework, representing the ``Generation with Retrieval" part of GRASP, allows a general LLM to gain domain knowledge by pulling relevant information from an external information source \cite{wang2024evaluatingqualityanswersretrievalaugmented}: in our case, this external source is the latest official set of municipal budget documents. This ensures that the LLM’s response is grounded in the factual data provided by those municipal budget documents, which in turn improves response truthfulness.

Our RAG framework was implemented with the following steps:
\begin{enumerate}
    \item Use LangChain’s PyPDFLoader class to chunk the budget documents
    \item Use OpenAI’s embeddings to convert the chunks to vectors
    \item Store the vectors in a ChromaDB vector database
    \item When a user asks a query, perform a similarity search against the vector database with the user query to get the matched vectors
    \item Send the matched vectors to the LLM for response generation
\end{enumerate}

\begin{figure*}[t]
    \centering
    \includegraphics[width=0.75\textwidth]{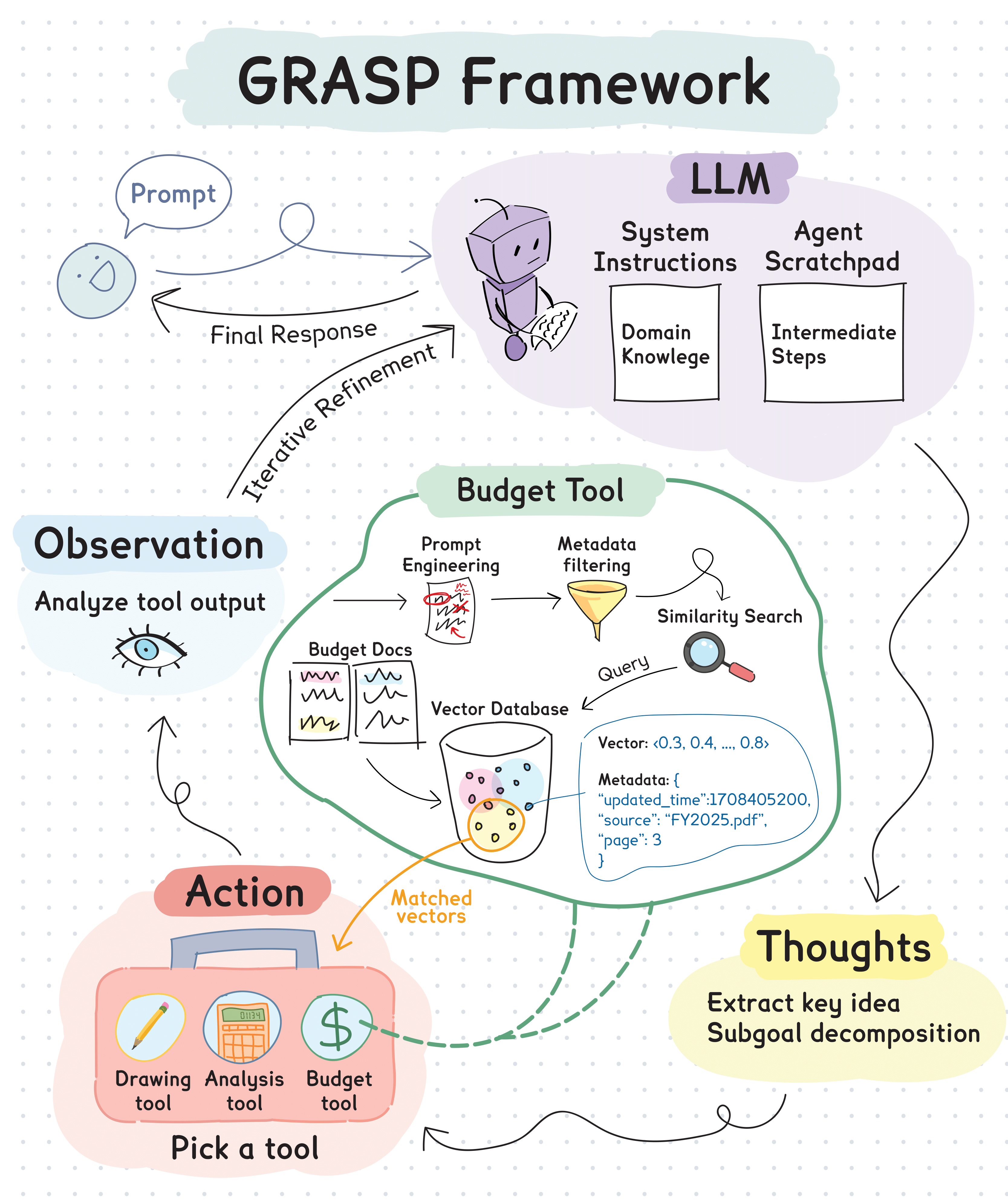}
    \caption{The GRASP framework combines two main frameworks: RAG and ReAct agent. The centerpiece titled “Budget Tool” describes the RAG framework, which allows for the usage of external documents as described in section III A. The outer circle describes the  thoughts-action-observation steps of a ReAct agentic workflow in section III B.}
    \label{fig:label1}
\end{figure*}

\subsection{Agentic Workflow (ReAct)}

The ReAct agent system, representing the ``Action System" part of GRASP, allows an LLM to decompose a user query into smaller, more manageable subtasks that individual tools (such as a budget tool or analysis tool) can achieve. These tools can also expand a chatbot's functionalities beyond what general LLMs can achieve: for example, a drawing tool could help a chatbot draw charts. Consider the query “draw a pie chart to represent how the total annual operating budget of the town is distributed”: the agent could first find the budget allocation by sector using a budget tool, and then draw the pie chart using a drawing tool.

The agentic workflow also gives the LLM the ability to reflect and critique its own responses, fetching more information using tools if it deems that it has insufficient knowledge to answer the query \cite{yao2022react}. This process of refinement occurs iteratively (unlike traditional LLMs which generate their response in one go), meaning the LLM will continue to revise the response until it is satisfied, hence improving truthfulness. Please see \Cref{fig:label1}.

We used LangChain’s Agent framework to implement the agentic workflow with the following steps:
\begin{enumerate}
    \item Create each tool as a derived class of the LangChain’s BaseTool class and implement each tool’s business logic in the run/arun methods
    \item Use LangChain’s ChatOpenAI to construct the LLM and bind it to the list of tools
    \item Setup the agent prompt template using LangChain’s ChatPromptTemplate class with the following components:
    \begin{itemize}
        \item LLM system instructions 
        \item Chat history (using LangChain’s MessagesPlaceholder class)
        \item Agent scratchpad, used by the agent to keep track of intermediate steps (also using LangChain’s MessagesPlaceholder class)
        \item User prompt
    \end{itemize}
    \item Instantiate LangChain’s AgentExecutor with the agent prompt template, LLM, and LangChain’s OpenAIToolsAgentOutputParser (which processes tool output). The AgentExecutor will be run each time a user passes in the query. 
\end{enumerate}

\subsection{Prompt Engineering}

We employ prompt engineering on two types of prompts:

\subsubsection{User prompt} due to the conversational nature of the interactions between a user and the chatbot, many queries must be implied based on the chat history. For example, if a user asks “What was the municipal school budget in FY2025?” followed by “What about the two years before?”, the chatbot would imply the second query actually means “What was the municipal school budget in FY2023 and FY2024?”. While the chatbot can understand these implied queries phrased exactly as the user typed them, such queries are usually lacking the keywords required for the similarity search performed by the RAG framework (in this case, the keywords would be school budget, FY2023, and FY2024). Hence, we instruct the LLM to rephrase queries to explicitly state all necessary information using the below prompt:

\begin{small}
\begin{verbatim}
Rephrase the following query to explicitly 
state the question and the year(s) in which 
the question is being asked: {currentQuery}

There may be additional context that is found 
in the previous user query: {lastQuery} 
\end{verbatim}  
\end{small}
    
We then use the rephrased query for similarity search to enhance the relevance of retrieved documents and thus truthfulness. 

\subsubsection{System prompt} we have been working with local officials to find domain-specific knowledge regarding municipal budget. We incorporate this domain knowledge into the LLM’s system prompt.

\subsection{Utilizing historical Data}

Since budget documents usually provide historical data,  an LLM trying to answer a question in the context of 2022 shouldn’t only look at the 2022 documents, but also the  documents from subsequent years like 2023, 2024, or 2025. In fact, due to the actual vs. projected distinction, it is often preferable to look in these later years. This domain-specific knowledge (that a general LLM would not be aware of) is incorporated into a GRASP chatbot as follows: 

First, determine which years the query is being applied to, which can be done by simply prompting the LLM. Once we have the list of years, modify it to include any subsequent years which could possibly contain the relevant historical data. From this modified list, construct a metadata filter that only allows the similarity search to retrieve documents from those years. 

\subsection{Grounding}

Our chatbot provides reference links to the exact pages in the original budget documents where it pulled information used in generating its response. This is done by chunking documents on a per-page basis, then associating each page's vector with “source” and “page” metadata fields.

\section{Code}

An example of a GRASP chatbot we have constructed for the town of Lexington, MA can be found here: \url{https://github.com/Brainana/LexBudget}. 

\section{Results}

\subsection{Interpretation of Results}
This experiment compared the effectiveness of three different LLMs—a GRASP chatbot, GPT 4o, and Google Gemini—across various town budget-related questions and categories, which includes questions about the general budget, revenues and expenditures, debt and deficits, and impact and outcome. This experiment also tested against questions relating to certain functionalities, which include table data retrieval questions, calculation questions, context-based questions, comparison over time questions, and sequential (follow-up) questions. To conduct this experiment, we manually ran each query through the three LLMs and compared their answer with the answer found in the official budget documents.

\begin{figure}[H]
    \centering
    \includegraphics[width=0.4\textwidth]{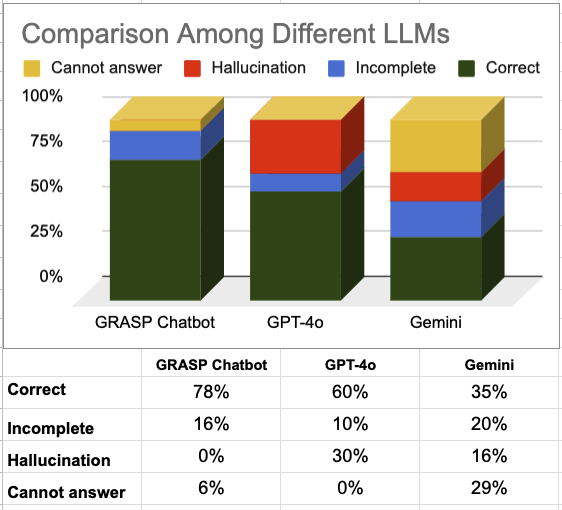}      
    \caption{Test results from running questions against 3 different AI tools.}
    \label{fig:label2}
\end{figure}

The findings of the experiment in Fig.~\ref{fig:label2} demonstrate that the GRASP chatbot consistently provided the most precise and accurate responses. While GPT 4o provided relatively accurate responses, they often lacked detail and depth. Gemini struggled the most, frequently failing to generate relevant data and offering less detailed information. Additionally, in terms of grounding, the GRASP chatbot's responses all provided citations linking directly to the exact page where the relevant information appears in the budget documents, while GPT-4o and Gemini only go as far as linking you to the documents. 

\subsection{Analysis in the Context of Existing Literature}
These results align with existing literature suggesting that hybrid models, such as those that utilize RAG and the ReAct agent paradigm, are more effective in producing accurate and comprehensive responses. RAG models can leverage external data sources, which enhances their capacity to provide relevant and detailed information, addressing one of the primary limitations that purely generative AI models face. The ReAct agent paradigm provides a framework that allows the LLM to perform iterative refinement rather than generating its responses in one go like traditional LLMs, and also allows the LLM to adapt to unforeseen circumstances such as the failure to retrieve desired information from a subtask. This supports the idea that integrating RAG and ReAct mechanisms with generative models improves overall accuracy and performance.

\subsection{Strengths and Limitations}
\subsubsection{Strengths}
Our experiment evaluated the chatbots across a wide range of questions and categories, providing a robust assessment of each model's performance. These assessments included not just the correctness but also the depth and relevance of the responses, offering a fine understanding of the capabilities of each chatbot.

\subsubsection{Limitations}
The questions, while diverse, may not encompass all potential real-world scenarios, limiting the ability to generalize the findings. Additionally, our experiment did not consider potential updates and improvements to the information retrieval capabilities of current LLMs, which have the potential to greatly improve the accuracy and reliability of their responses.

\section{Conclusion}

Creating an AI chatbot to answer questions about the municipal budget makes it much easier for residents or the general public to reliably and efficiently acquire accurate, comprehensible budget information about their local government. However, a significant amount of effort needs to be put in to minimize LLM hallucinations in such applications.  As such, we proposed GRASP, a framework combining the RAG and ReAct Agent frameworks, prompt engineering, and domain knowledge. Furthermore, we proposed refining the responses and domain knowledge by working with domain experts from local governments. Such a system would make finding budget information more reliable and convenient for people of all levels of familiarity with their town’s financial workings, from town officials to ordinary citizens. In addition, the methodology of GRASP makes it easy for this system to be generalized for all towns and cities, which could even allow such systems to perform inter-municipal comparisons.

Providing residents with accurate and comprehensive information is the key to enhancing community engagement, empowering citizens with a better understanding of their town budget while reserving the right to make decisions and judgments with the people.

\section*{Acknowledgments}

We would like to express our sincere gratitude to Prof. Wei Ding and Prof. Ping Chen for their insightful guidance in shaping the direction and quality of this research paper. 

We would like to thank our mentors Andrei Radulescu-Banu, Jeannie Lu, Nagarjuna Venna, Chester Curme, Prof. Wei Ding, Neerja Bajaj, and John Truelove for the invaluable mentorship, encouragement and support throughout the chatbot’s development and for their help in promoting the project.

We would like to thank  Kevin Zhu, Cassidy Xu, Emma He, William Yang, and Andrew Pan for their contribution to the creation of the chatbot.

We would like to thank the Lexington town meeting members Glenn Parker and Meg Muckenhoupt for sharing their domain knowledge with regards to the town budget and providing crucial feedback on our chatbot.

We would like to thank the LexObserver organization and their founder Nicco Mele for providing the funding for the research and development of the GRASP chatbot.

Lastly, we would like to thank the Lexington STEAM Team organization which we are a part of for continuing to provide opportunities for high school students to do impactful community service projects.

\bibliography{main}

\end{document}